\def\paperID{0000}
\def\confName{CVPR}
\def\confYear{2024}

\def\authorBlock{
    Xinli Guo, 
    Weidong Zhang$^{*}$, 
    Ruonan Liu, 
    Peng Han,
    and Hongtian Chen 
    
}

\newif\ifreview 
\newif\ifarxiv 
\newif\ifcamera 
\newif\ifrebuttal 

\documentclass[conference]{IEEEtran}
\IEEEoverridecommandlockouts
\usepackage{graphicx}	
\usepackage{amsmath}	
\usepackage{amssymb}	
\usepackage{booktabs}
\usepackage{times}
\usepackage{microtype}
\usepackage{epsfig}
\usepackage[table,xcdraw,dvipsnames]{xcolor}
\usepackage{caption}
\usepackage{float}
\usepackage{cite}
\usepackage{placeins}
\usepackage{color, colortbl}
\usepackage{stfloats}
\usepackage{enumitem}
\usepackage{tabularx}
\usepackage{xstring}
\usepackage{multirow}
\usepackage{xspace}
\usepackage{url}
\usepackage{subcaption}
\usepackage{xcolor}
\usepackage[hang,flushmargin]{footmisc}

\definecolor{cvprblue}{rgb}{0.21,0.49,0.74}
\usepackage[pagebackref,breaklinks,colorlinks,citecolor=cvprblue]{hyperref}
\usepackage{cleveref}
\ifcamera \usepackage[accsupp]{axessibility} \fi

% ...

\ifarxiv  \fi

\newcommand{\R}[1]{{%
    \textbf{%
        \ifstrequal{#1}{1}{\textcolor{red}{R#1}}{%
        \ifstrequal{#1}{2}{\textcolor{blue}{R#1}}{%
        \ifstrequal{#1}{3}{\textcolor{magenta}{R#1}}{%
        \ifstrequal{#1}{4}{\textcolor{teal}{R#1}}{%
                           \textcolor{cyan}{R#1}%
        }}}}%
    }%
}}

\begin{document}
\title{MotionGS : Compact Gaussian Splatting SLAM by Motion Filter}


\author{\authorBlock\thanks{\quad All authors are with Shanghai Jiao Tong University. $^{\ast}$Corresponding Author: Weidong Zhang (wdzhang@sjtu.edu.cn).}
\thanks{\quad This paper is partly supported by the National Key R\&D Program of China (2022ZD0119900), the National Natural Science Foundation of China (U2141234), Shanghai Science and Technology program (22015810300), Hainan Province Science and Technology Special Fund (ZDYF2024GXJS003), National Natural Science Foundation of China (62303308),  Shanghai Pujiang Program (23PJ1404700), Joint Research Fund of Shanghai Academy of Spaceflight Technology (USCAST2023-22).}}

\maketitle


\begin{figure*}[tp]
    
    \centering
    \includegraphics[width=\textwidth]{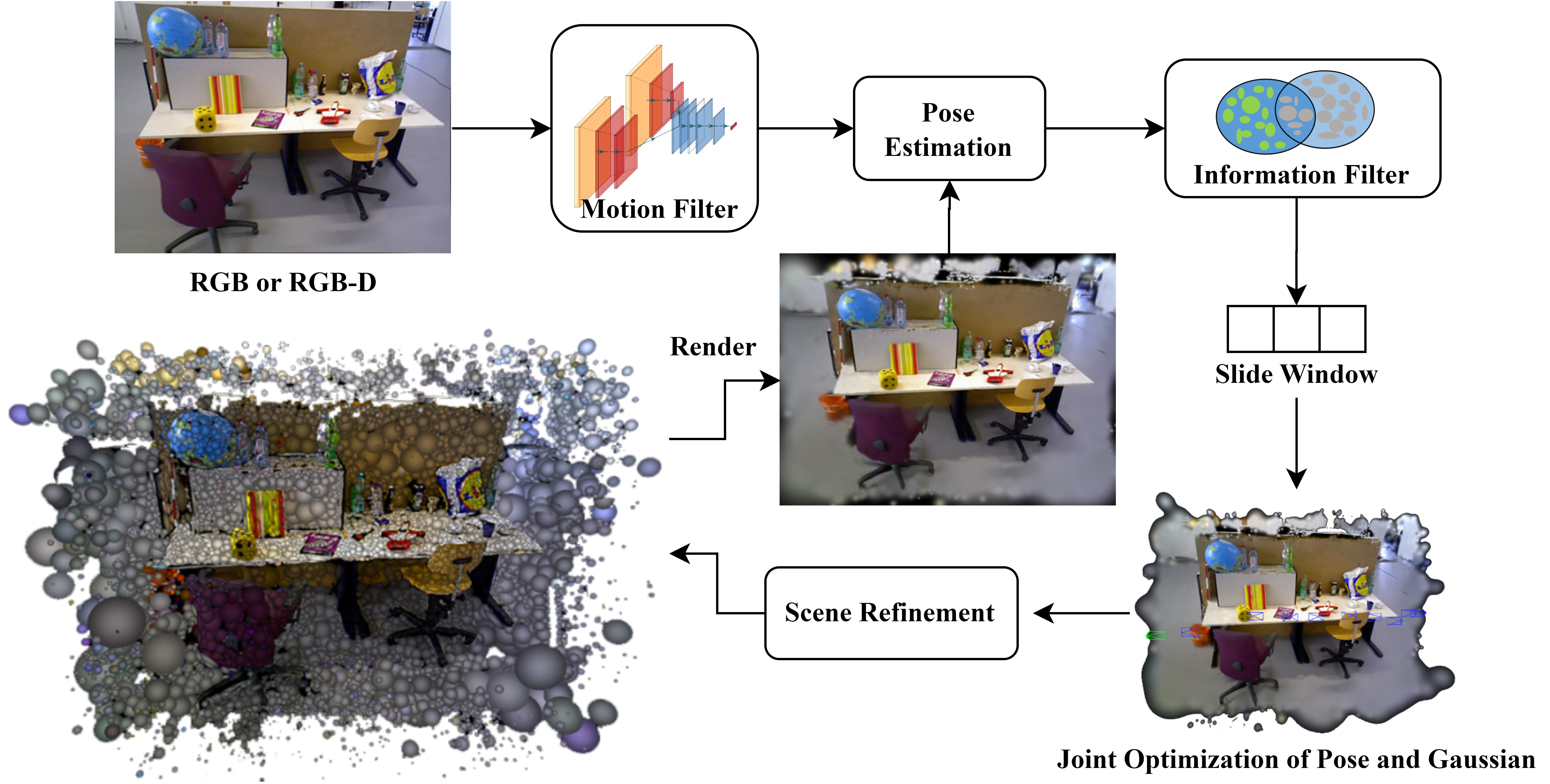}
    \caption{\textbf{Overview of MotionGS.} The input to MotionGS at each timestep is the current RGB-D/RGB image. After the motion filter, the directly pose optimization of the motion keyframe is done based on the photometric error between the GT and render result. After information filter, the joint optimization of keyframe poses and 3D scene geometry on sliding windows and random historical frames is carried out in the mapping thread. Finally, the scene is refined.}
    
    \label{fig:framework}
\end{figure*}

\begin{abstract}
With their high-fidelity scene representation capability, the attention of SLAM field is deeply attracted by the Neural Radiation Field (NeRF) and 3D Gaussian Splatting (3DGS). Recently, there has been a surge in NeRF-based SLAM, while 3DGS-based SLAM is sparse. A novel 3DGS-based SLAM approach with a fusion of deep visual feature, dual keyframe selection and 3DGS is presented in this paper. Compared with the existing methods, the proposed tracking is achieved by feature extraction and motion filter on each frame. The joint optimization of poses and 3D Gaussians runs through the entire mapping process. Additionally, the coarse-to-fine pose estimation and compact Gaussian scene representation are implemented by dual keyframe selection and novel loss functions.  Experimental results demonstrate that the proposed algorithm not only outperforms the existing methods in tracking and mapping, but also has less memory usage.
\end{abstract}
\section{Introduction}
\label{sec:intro}
The main purpose of simultaneous localization and mapping (SLAM) is to enable real-time positioning and simultaneous mapping in unknown environments. This technology plays a crucial role in various fields such as autonomous driving \cite{drive}, unmanned systems \cite{water}, virtual reality \cite{VR}, and augmented reality \cite{AR}. 
Over the past years, the SLAM field has made remarkable strides in both localization accuracy and speed \cite{orbslam3}. However, recent attention has been attracted towards scene representation and comprehension. This shift has given rise to numerous map-centric SLAM. The scene representation of traditional map-centric SLAM includes point clouds or surfels \cite{ORB-SLAM2,surfel,surfel2,surfel3} , meshs \cite{mesh,mesh2}, voxels \cite{voxel,voxel1,voxel2,voxels}.  As a pivotal component of SLAM systems, the selection of the mapping approach not only shapes the overall system design but also defines its capabilities.\par
In term of dense visual SLAM, these classic map-centric methods can not achieve the high-fidelity representation nor fail in reconstructing the fine textures and repetitive scenes. 
Recently, Neural Radiance Fields (NeRF) \cite{nerf} have garnered significant attention in the SLAM field. NeRF represents a novel view synthesis approach with an implicit scene representation, which has the advantages of photorealism and minimal memory usage. A multitude of NeRF-based dense SLAM methods has sprung up with respective to this evolving NeRF \cite{nerf_imap,vox-fusion,niceslam,Eslam,goslam,coslam,pointslam,nerfslam}. However, NeRF-based dense SLAM methods all rely on ray-based volume rendering by dense sampling, which is highly time-consuming and unreliable \cite{gsslam}.\par
 As a new scene representation, 3D Gaussian Splatting (3DGS) \cite{3dgs} explicitly represents scenes using 3D Gaussian as primitives, achieving rendering effects comparable to NeRF while offering faster speed of optimization and rendering. 3DGS is exceptionally well-suited as the mapping method in SLAM systems, as demonstrated by the  recently proposed approaches \cite{gsslam,monogs,splatam}. \par
 In this paper, a novel dense 3DGS-based SLAM approach, named MotionGS, is proposed. Deep feature extraction, dual keyframe selection, and 3DGS are integrated in this method. In the tracking thread, we extract deep features from each image, obtain motion keyframes through motion filter, and employ direct pose optimization tailored for 3DGS to reduce the number of processed frames and improve tracking performance. Finally, the tracking thread continuously updates and maintains a sliding window of information keyframes by information filter. In the mapping thread, a new loss function is designed to simultaneously optimize keyframe pose and 3D Gaussian in the diff-gaussian-rasterization framework. The dual optimizations in the tracking and mapping threads achieve coarse-to-fine pose optimization and reduce storage needs. Extensive evaluations on indoor RGB-D datasets have demonstrated the state-of-the-art performance of our approach in tracking,  rendering, and mapping. In summary, the main contributions are given as follows:
\begin{enumerate}
    \item A novel dense visual SLAM method based on 3DGS (that integrates deep feature extraction, dual keyframe selection and 3DGS) is proposed. This approach not only achieves accurate real-time tracking and high-fidelity reconstruction, but also supports both RGB and RGB-D inputs.
    \item A novel dual keyframe strategy and a new loss function are designed to enhance the tracking accuracy and maintain rendering quality while reducing memory consumption.
    \item Our approach achieves state-of-the-art performance on Replica and TUM-RGBD datasets in terms of tracking, mapping and runs at 2.5 fps with less memory usage.
\end{enumerate}
\section{Related Work}
\label{sec:related}
\textbf{Dense visual SLAM}: Traditional dense visual SLAM mainly consists of two types: stereo vision and RGB-D dense mapping. For example, \textbf{DTAM} \cite{DTAM} achieved real-time camera pose tracking and point cloud based mapping by estimating optical flow and depth between consecutive frames, while \textbf{ORB-SLAM2 pro} achieved real-time pose tracking and point cloud based mapping in RGB-D mode using conventional feature point methods.  \textbf{ElasticFusion} \cite{surfel3} achieved elastic registration and nonlinear optimization for real-time pose estimation and map reconstruction. \textbf{Kintinuous} \cite{Kintinuous} combined dense visual SLAM with RGB-D sensors and motion estimation, enabling long-time high-quality  map reconstruction. Traditional dense mapping methods include point clouds, triangulated meshes, surfels, and voxels. Recently, the fusion of deep learning with traditional SLAM frameworks has achieved more precise localization and reconstruction effects. For example, \textbf{DROID-SLAM} \cite{droid} extracted dense feature maps and iteratively updated camera poses and pixel-level depth by the depth update operators. It achieved excellent real-time positioning and dense map reconstruction.\\
\textbf{NeRF-based SLAM}: According to sensor types, this category can be divided into two forms: RGB and RGB-D. \textbf{iMAP} \cite{nerf_imap} was the first RGB-D SLAM method to adopt implicit neural representation. The idea of using MLP for scene representation had been widely recognized and promoted. \textbf{NICE-SLAM} \cite{niceslam} employed a hierarchical strategy by using multiple MLPs to encode scene geometry into (coarse, medium, and fine) three voxel grids. The centralized three-layer voxel grid significantly improved the precision and efficiency of the optimization process and effectively replied catastrophic forgetting issues. Furthermore, \textbf{ESLAM} \cite{Eslam} adopted large-scale axis-aligned representations of three coarse and three fine feature planes with TSDF as the scene geometry representation. This greatly reduced memory usage and significantly improved frame rates. For the same purpose, \textbf{Co-SLAM} \cite{coslam} and \textbf{Point-SLAM} \cite{pointslam} respectively utilized encoding of joint coordinate with  parameter and neural point cloud representation to enhance mapping performance. As for RGB forms,  based on the DROID-SLAM algorithm, \textbf{Hi-SLAM} \cite{hislam} supplemented with depth priors and joint depth and scale adjustment to compensate for the lack of depth information.\\
\textbf{3DGS-based SLAM}: 
Research in this field is rapidly developing, but there are relatively few related algorithms. In addition to the differentiable optimization and fast re-rendering of 3DGS, \textbf{GS-SLAM} \cite{gsslam} used adaptive expansion strategy and coarse-to-fine camera tracking policy to achieve high fidelity and fast high-resolution rendering performance. \textbf{MonoGS} \cite{monogs} adopted a direct optimization approach against 3D Gaussian ellipsoid for camera tracking, coupled with geometric verification and regularization techniques, achieving continuous high-fidelity 3D scene reconstruction of even small and transparent objects. \textbf{SplaTAM} \cite{splatam} employed a customized online tracking and mapping system, combined with silhouette masks to achieve fast rendering, dense optimization, efficient map reconstruction, and structured map expansion. \textbf{Photo-SLAM} \cite{photoslam} integrated explicit geometric features and implicit texture representations into a super-primitive map based on the classic ORB-SLAM3 method.  It also introduced a training method based on Gaussian pyramid to gradually learn multi-level features, improving the performance of realistic mapping.  \textbf{Gaussian-SLAM} \cite{gaussianslam} introduced several new techniques such as differential depth rendering, explicit Gaussian gradients, and frame-to-model alignment, significantly enhancing its performance. However, recent methods still exhibited slightly inferior performance in frame processing rates and memory usage. \textbf{GS-ICP-SLAM} \cite{gsicp} took a different approach by utilizing point cloud matching to greatly increase frame processing rates.

\section{The Proposed Approach}
\label{sec:method}
\subsection{Compact 3DGS Scene Representation}
\label{subsec:Compact 3DGS Scene Representation}
3DGS explicitly represents the scene with a set of anisotropic Gaussian. Each Gaussian $g$ encapsulates properties: $c_g$, $o_g$, $sh_g$, $p_g$, and $\Sigma_g$, representing color, opacity, spherical harmonics, position in the world coordinate system, and Gaussian covariance matrix, respectively. The explicit voxel-based rendering of 3DGS representation can render the color $c_{p}$ and depth $d_{p}$ of specified pixels by frustum culling and blending $N$ Gaussian:\\
\begin{gather}
     c_{p} = \sum_{i}^{N} c_{i} o_{i} \prod_{j}^{i-1} (1-o_{j} ),\\
     d_{p} = \sum_{i}^{N} d_{i} o_{i} \prod_{j}^{i-1} (1-o_{j} ),
\end{gather}

\noindent where $c_{i}$ represents the color of the $i$th Gaussian along with the camera ray; $d_{i}$ represents the distance from the camera along with the ray to the $i$th Gaussian. In the process of splatting in 3DGS, the differentiable transformation function from 3D-Gaussian $(p_{w},\Sigma_{w})$ to 2D-Gaussian $(p_{c},\Sigma _{c})$ is expressed as follows:
\begin{equation}
    p_{c} = T_{cw} \ast p_{w} , \Sigma _{c} = JW\Sigma _{w} W^{T} J^{T},
\end{equation}
where $c$ and $w$ respectively represent camera and world Coordinate System; $T_{cw}\in SE^{3}$ represents camera poses; $J$ is the linear approximation of the Jacobian matrix for the projection transformation; $W$ is the rotational component of $T_{cw}$.
Given that the Gaussian geometry in the scene is similar to an ellipsoid, its covariance matrix $\Sigma$ can be represented in an ellipsoidal form:
\begin{equation}
    \Sigma = RSS^{T}R^{T},
\end{equation}
where $R$ and $S$ represent the rotation and scaling matrix of the Gaussian ellipsoid derived from quaternion $q$ and scale $s$, respectively.
Based on the above formula, the open-source diff-gaussian-rasterization \cite{3dgs} achieves differentiable 3D Gaussian rasterization.  This library is founded on differentiable rendering gradients, streamlining computations in void spaces and thereby expediting both training and rendering processes. In each iteration, the GS scene is optimized end to end by the weighted sum of $L_{1}$ and structural similarity index ($L_{ssim}$) between the groundtruth and rendered image.\par
Inspired by \cite{compact}, a new penalty term is set based on the scale and opacity of Gaussian in the optimization process to compactly represent the scene while preserving original performance. The binary masks $B\in[0,1]$ are set on the opacity $o_{i}$ and scale $s_{i}$ of each Gaussian and participate in the optimization process. To make them differentiable\cite{estimating}, an auxiliary parameters $b\in{R}$ is introduced. The masked scale $\hat{s} _{i}$ and masked opacity $\hat{o} _{i}$ are given as follows:
\begin{gather}
    B_{i} = Stop(1[\sigma (b_{i})< \epsilon ]-\sigma(b_{i})) + \sigma(b_{i}), \\
    \hat{s} _{i} = B_{i} s_{i} ,\hat{o} _{i} = B_{i} o_{i} ,
\end{gather}
\noindent where $Stop$ and $\sigma$ respectively represent the stop gradient and sigmoid function. This mask eliminates the occlusion of Mini-Gaussian, which has little effect during rasterization rendering. In each densification, besides splitting, cloning, and pruning, Based on this mask, some scene geometries are also removed to reduce the number of Gaussian involved in optimization. Moreover, to balance high-fidelity rendering with reducing scene geometries, a new mask loss is designed:\\
\begin{equation}
    L_{M} = \frac{1}{N}\sum_{i=1}^{N}\sigma(b_{i}).
\end{equation}
 Thus, the complete loss function is constructed as follows:
\begin{equation}
    L =(1-\lambda_{1})L_{1}+\lambda_{1}L_{ssim}+\lambda_{2}L_{M} 
    \label{eq:new loss},
\end{equation}
where $\lambda_{1}$ and $\lambda_{2}$ represent the weights. After optimization based on this loss function (\ref{eq:new loss}) is finished, the masks will be discarded as there are no longer useful.
\subsection{Dual Keyframe Strategy}
The dual keyframe strategy consists of a motion filter and an information filter, respectively corresponding to motion keyframe and information keyframe.\par

\textbf{Motion Filter} It performs feature extraction on each frame and only retains frames that exceed the correlation threshold. Similar to \cite{droid}, three different networks are designed as: the feature network, context network, and update network. \par
Each input image is processed by the feature network consisting of six residual modules and three downsampling modules, producing dense feature maps at 1/8 resolution of the original image size. Given that the input image $img \in R^{H\times W \times 3}$, the output dense feature map.The ouput dense feature map $dmap \in R^{H/8\times W/8 \times D}$ where we set $D = 256$. These dense feature maps are used to construct correlation pyramids and obtain sets of correlated feature vectors, while finer texture features extracted by the context network are added to the update network during each update process. Inter-frame motion vector is obtained after the update process.\par
\textbf{Correlation Pyramid} For any pair of frames $(I_{i},I_{j})$ in the frame set, the 4D correlation vectors $Cor \in R^{}$ are computed by taking the dot product of all pairs of feature vectors from the dense feature maps $dmap_{i}$ and $dmap_{j}$.
\begin{equation}
    Cor(I_{i},I_{j})_{u_{1}v_{1}u_{2}v_{2}}=\left \langle dmap(I_{i})_{u_{1}v_{1}},  dmap(I_{j})_{u_{2}v_{2}}\right \rangle.
\end{equation}
Then, average pooling is performed on the last two dimensions of the relevant vectors to construct a 4-level correlation pyramid. We also employ a query operation, which takes an $H\times W$ coordinate grid as the input and retrieves values from the correlation volume using bilinear interpolation. This operator is applied to each correlation volume in the pyramid, and the final feature vector is computed by concatenating the results from each level.\par
During the motion filter process, the motion threshold and maximum frame interval are preset. When the norm of its inter-frame motion vectors exceeds the motion threshold or its inter-frame interval with respect to the previous motion keyframe exceeds the maximum frame interval, the current frame is added as a new motion keyframe. The initial pose of the new motion keyframe is set based on the motion model, where the motion vector is defined as the pose transformation between the previous motion keyframe and the one before it.\par
\textbf{Information Filter}
Considered the high real-time demand of SLAM algorithms,  it is impractical to use all motion keyframes for optimization.
Therefore, the tracking thread performs dual keyframe selection to update and maintain a sliding window of information keyframes. The information threshold and minimum mapping frame distance are preset. A motion keyframe is added to the sliding window if its relative complement to the previous information keyframe exceeds the threshold or its distance to the previous informative keyframe exceeds the minimum mapping frame distance. To maintain a constant window size, deletion operation is performed as well. An information keyframe is removed if its overlap coefficient with the new informative keyframe falls below a threshold or its inter-frame displacement from the previous informative keyframe is minimal.
\par

For the motion keyframe $i$, its relative complement ($RC$) and overlap coefficient ($OC$) are defined as follows:
\begin{equation}
    RC(i,j)=\frac{G_{i}-G_{i}\cap G_{j} }{G_{i}\cup G_{j}},
\end{equation}
\begin{equation}
    OC_{i,j}=\frac{G_{i}\cap G_{j} }{min(G_{i},G_{j})},
\end{equation}
where $G_{i}$ represents the scene geometry in the $i$th frame; $j$ denotes the preceding information keyframe for motion keyframe $i$, while $k$ denotes the preceding information keyframe for information keyframe $j$.
\subsection{3DGS-based Direct pose optimization}
Firstly, it is crucial to acknowledge that the rendering results of 3DGS inherently exhibit some degree of blur. It is impractical to optimize the partial pixels between real photos and rendered images. Although the pose optimization using the entire photo may slightly lag in speed, it aligns with the differentiable rendering framework of 3DGS. Therefore, all pixels are directly used to the pose optimization framework based on the photometric error between real photos and rendered images. The photometric error relative to the camera pose and depth (if available) for pose optimization is defined as follows:
\begin{gather}
    E_{rgb} =opacity * \left \| I_{render}-I_{gt} \right \| _{1},\\
    E_{d} = opacity * \left \| D_{render}-D_{gt} \right \| _{1}.
    \label{eq:9}
\end{gather}
Inspired by \cite{monogs}, the Lie algebra helps find the smallest Jacobian matrix with the right number of degrees of freedom, so no extra calculations are needed.The chain expansion of the transformation function for rasterization rendering in 3DGS is expressed as follows:
\begin{equation}
    \frac{\partial p_{c}}{\partial T_{cw}} =\frac{\partial p_{c} }{\partial p_{w} }\frac{\partial p_{w}}{\partial T_{cw}},
\end{equation}
\begin{equation}
    \frac{\partial \Sigma_{c}}{\partial T_{cw}} =\frac{\partial \Sigma_{c} }{\partial J }\frac{\partial J}{\partial p_{c}}\frac{\partial p_{c}}{\partial T_{cw}}+
\frac{\partial \Sigma_{c}}{\partial W}\frac{\partial W}{\partial T_{cw}}.
\end{equation}
The Jacobian matrix of a function $f$ acting on a manifold was proposed in \cite{micro}:
\begin{equation}
    \frac{^{\chi }\partial f(\chi ) }{\partial\tau} \overset{\bigtriangleup }{=} \lim_{\tau  \to 0}\frac{\partial
 Log(f(\chi )^{-1}\circ  f(\chi \circ Exp(\chi )))}{\partial \tau } , 
\end{equation}
where $\chi \in SE^{3}$; $\tau \in \mathfrak{se}^{3}$; $\circ$ denotes the group operator; $Exp$ and $Log$ respectively represent the exponential and logarithmic functions mapping from the Lie algebra to the Lie group. Therefore, we can get:
\begin{equation}
    \frac{\partial p_{c} }{\partial T_{cw}}=[1,-p_{c}^{\times}],
\end{equation}
\begin{equation}
    \frac{\partial W }{\partial T_{cw}}=
\begin{bmatrix}
 0 & W_{:,1}^{\times}\\
 0 & W_{:,2}^{\times}\\
 0 & W_{:,3}^{\times}
\end{bmatrix},
\end{equation}
where $\times$ represents the skew-symmetric matrix of a three-dimensional vector.
\subsection{SLAM System}
\label{sec:slamsystem}
Figure \ref{fig:framework} illustrates the overall framework of the proposed MotionGS. Integrating the aforementioned modules, MotionGS mainly consists of two parts: tracking and Gaussian dense mapping. Each part runs as an independent thread and maintains communication with each other, collectively constructing the high-fidelity dense  map reconstruction .
\subsubsection{Tracking}
The tracking thread is responsible for the dual keyframe strategy: motion and information keyframes. Motion keyframes are primarily utilized for tracking purposes, whereas information keyframes are crucial for mapping.\par
The tracking thread performs feature extraction and motion filter on each frame image. It sets the initial pose of the current motion keyframe based on the previous motion keyframe and then minimizes the photometric error between the groundtruth image and the rendered image to obtain a coarse pose of the motion keyframe. Finally, the tracking thread continuously updates and maintains a sliding window of information keyframes. 
\subsubsection{Mapping}
The mapping thread is responsible for the joint optimization of 3D scene geometries and keyframe poses, as well as real-time rendering. As the main content of interaction between tracking and mapping, the information keyframe sliding window is directly used for the joint optimization. 
The Adam optimizer in pytorch is utilized and the mapping loss $L_{map}$ between rendered images and groundtruth images is showed below.
\begin{gather}
    L_{map}=\sum_{i}^{m+n}(\left \| I_{i}^{*}-I^{gt}_{i}\right \|_{1}) +\beta \sum_{j}^{G}\left \| S_{j} - \bar S\right \|_{1}),
\end{gather}
where $m$ represents the size of the sliding window, which is the number of frames used to compute the photometric loss within the window; $n$ denotes the number of randomly selected historical frames; $I^{*}$ denotes the rendered results; $G_{i}$ represents the set of all Gaussians; $S$ denotes the scale of Gaussian and the $\bar S$ denotes the mean scale of all Gaussians.\par
In addition to general splitting and pruning, the color refinement of the scene will also be regularly performed based on the new loss function (\ref{eq:new loss}). After the tracking is completed,  the entire scene refinement is done in the mapping thread like SfM. In scene refinement, redundant Gaussian are removed, and the size of the scene model is reduced accordingly.

\section{Experiment}
\label{sec:experiment}
\subsection{Experimental Setting}
The experiments have been conducted on two well-known datasets, TUM RGB-D \cite{tum} (3 sequences)  and Replica \cite{replica} (8 sequences), to compare the performance of the MotionGS with other methods. The algorithm results are obtained on the server computer equipped with 64 cores and RTX 3090 GPU.
\subsubsection{Metrics}
As the two key components of SLAM, tracking and mapping are both crucial. For the tracking part, the root mean square error (RMSE) of the absolute trajectory error (ATE) of keyframes is calculated based on the tracking result of each algorithm. For the mapping part, the standard photometric rendering quality metrics are used, including: the peak signal to noise ratio (PSNR), structural similarity index (SSIM), and learned perceptual image patch similarity (LPIPS). Additionally, the storage usage for map representation  is also analyzed and compared . 
\subsubsection{Baseline Methods}
We compare and analyze MotionGS against classic traditional SLAM approach (ORB-SLAM2 \cite{ORB-SLAM2}), deep learning based method (DROID-SLAM \cite{droid}), NeRF-based SLAM methods (iMAP \cite{nerf_imap}, NICE-SLAM \cite{niceslam}, Vox-Fusion \cite{vox-fusion}, ESLAM \cite{Eslam}, Co-SLAM \cite{coslam}, Point-SLAM \cite{pointslam}), and 3DGS-based SLAM methods (MonoGS \cite{monogs}, SplaTAM \cite{splatam}, GS-SLAM \cite{gsslam}). For non-dense SLAM methods, only tracking accuracy is compared. The metrics of the baselines are took from 
 \cite{monogs,splatam,gsslam}.
\subsection{Evaluation}
\subsubsection{Track} 

Table \ref{tab:ate} displays the ATE metrics of all above methods on the TUM dataset. In the RGB-D setting, MotionGS has surpassed the state-of-the-art methods of the 3DGS-based SLAM and NeRF-based SLAM in the fr1 and fr3 scenes, only slightly inferior to the baseline methods in the fr2 scene. In the Mono setting, MotionGS outperforms MonoGS, which is also applicable to monocular setups. However, MotionGS still exhibits higher errors compared with DROID-SLAM and ORB-SLAM2 due to the absence of loop closure detection and global BA. This demonstrates the pivotal role of loop closure detection and global BA in enhancing the localization accuracy of SLAM methods and their potential in 3DGS scene.

Table \ref{tab:ate2} presents the ATE metrics of all above methods on the Replica dataset. While the baseline methods have achieved remarkable millimeter-level accuracy, MotionGS remains competitive, surpassing them in five out of eight scenes. In the o0 and r0 scenarios, the best method only outperforms MotionGS by 0.14 cm in ATE.
The excellent performance shown in Tables \ref{tab:ate} and \ref{tab:ate2} is mainly attributed to the coarse to fine pose estimation. Tracking thread makes a coarse pose optimizition of the motion keyframe to reduce the accumulated tracking error caused by pose estimation for consecutive frames. Then,  the joint optimization of poses and 3D Gaussians  in Mapping thread further improves the tracking accuracy.

\begin{table}[htbp]
\centering
\scalebox{1}{

\begin{tabular}{cccccc}
\toprule
\textbf{Class} & \textbf{Method} & \textbf{fr1} & \textbf{fr2} & \textbf{fr3} & \textbf{Arg}\\
\midrule
\multirow{4}*{\textbf{Mono}} &  ORB-SLAM2 & 2.00 & 0.60 & 2.30 & 1.60 \\

~ & DROID-SLAM & 1.80 & 0.50 & 2.80 & 1.70 \\
~ & MonoGS  & 4.15 & 4.79 & 4.39 & 4.44 \\
~ & \textbf{Ours}      & \textbf{3.53} & \textbf{3.93} & \textbf{2.43} & \textbf{3.30} \\
\midrule
\multirow{6}*{\textbf{RGB-D(NeRF)}}&iMAP       & 4.90 & 2.00 & 5.80 & 4.23 \\
~ & NICE-SLAM  & 4.26 & 6.19 & 6.87 & 5.77 \\
~ & ESLAM      & 2.47 & \textcolor{red}{\textbf{1.11}} & 2.42 & \textcolor[HTML]{009901}{\textbf{2.00}} \\
~ & Vox-Fusion & 3.52 & 1.49 & 26.01 & 10.34 \\
~ & Co-SLAM    & \textcolor[HTML]{009901}{\textbf{2.40}} & 1.70 & \textcolor[HTML]{009901}{\textbf{2.40}} & 2.17 \\
~ & Point-SLAM & 4.34 & \textcolor[HTML]{009901}{\textbf{1.31}} & 3.48 & 3.04 \\
\midrule
\multirow{4}*{\textbf{RGB-D(3DGS)}} & MonoGS     & \textcolor{blue}{\textbf{1.52}} & 1.58 & \textcolor{blue}{\textbf{1.65}} & \textcolor{blue}{\textbf{1.58}} \\
~ & SplaTAM    & 3.35 & \textcolor{blue}{\textbf{1.24}} & 5.16 & 3.25 \\
~ & GS-SLAM & 3.30 & 1.32 &  6.60 &  3.70 \\
~ & \textbf{Ours}      & \textcolor{red}{\textbf{1.47}} & \textbf{1.38} & \textcolor{red}{\textbf{1.41}} & \textcolor{red}{\textbf{1.46}} \\
\bottomrule
\end{tabular}

}
\captionsetup{labelfont=bf}
\caption{
\textbf{Comparison of tracking results ATE (cm) on TUM}. The red, blue, and green in the above table represent the first, second, and third, respectively.
}
\vspace{-0.05in}
\label{tab:ate}
\end{table}
\begin{table}[htbp]
\centering
\scalebox{0.85}{%

\begin{tabular}{cccccccccc}
\toprule
\textbf{Method} & \textbf{r0} & \textbf{r1} & \textbf{r2} & \textbf{o0} & \textbf{o1} & \textbf{o2} & \textbf{o3} & \textbf{o4} & \textbf{Arg} \\
\midrule
iMAP       & 3.12 & 2.54 & 2.31 & 1.69 & 1.03 & 3.99 & 4.05 & 1.93 & 2.58\\
NICE-SLAM  & 0.97 & 1.31 & 1.07 & 0.88 & 1.00 & 1.06 & 1.10 & 1.13 & 1.07\\
Vox-Fusion & 1.37 & 4.70 & 1.47 & 8.48 & 2.04 & 2.58 & 1.11 & 2.94 & 3.09 \\
ESLAM      & 0.71 & 0.70 & 0.52 & 0.57 & 0.55 & 0.58 & 0.72 & \textcolor{blue}{\textbf{0.63}} & 0.63 \\
Point-SLAM & 0.61 & \textcolor[HTML]{009901}{\textbf{0.41}} & \textcolor[HTML]{009901}{\textbf{0.37}} & \textcolor{blue}{\textbf{0.38}} & 0.48 & 0.54 & 0.69 & 0.72 & \textcolor[HTML]{009901}{\textbf{0.53}}\\
MonoGS     & \textcolor[HTML]{009901}{\textbf{0.47}} & 0.43 & \textcolor{red}{\textbf{0.31}} & 0.70 & 0.57 & \textcolor[HTML]{009901}{\textbf{0.31}} & \textcolor[HTML]{009901}{\textbf{0.31}} & 3.2 & 0.79 \\
SplaTAM    & \textcolor{red}{\textbf{0.36}} & \textcolor{blue}{\textbf{0.31}} & 0.40 & \textcolor{red}{\textbf{0.29}} & \textcolor[HTML]{009901}{\textbf{0.47}} & \textcolor{blue}{\textbf{0.27}} &\textcolor{blue}{\textbf{ 0.29}} & \textcolor{red}{\textbf{0.32}} & 0.55 \\
GS-SLAM & 0.48 & 0.53 &\textcolor{blue}{\textbf{0.33}}& 0.52& \textcolor{blue}{\textbf{0.41}} &0.59& 0.46& \textcolor[HTML]{009901}{\textbf{0.7}}& \textcolor{blue}{\textbf{0.50}}\\

\textbf{Our}        & \textcolor{blue}{\textbf{0.46}} & \textcolor{red}{\textbf{0.28}} & \textcolor{red}{\textbf{0.31}} & \textcolor[HTML]{009901}{\textbf{0.43}} & \textcolor{red}{\textbf{0.27}} & \textcolor{red}{\textbf{0.19}} & \textcolor{red}{\textbf{0.14}} & \textbf{1.85} & \textcolor{red}{\textbf{0.49}} \\
\bottomrule
\end{tabular}

}
\captionsetup{labelfont=bf}
\caption{\textbf{Comparison of tracking results ATE (cm) on Replica.} The red, blue, and green in the above table represent the first, second, and third, respectively.
}
\vspace{-0.05in}
\label{tab:ate2}
\end{table}

\subsubsection{Render} 
\begin{figure*}[htbp]
    \centering
    \includegraphics[width=\textwidth]{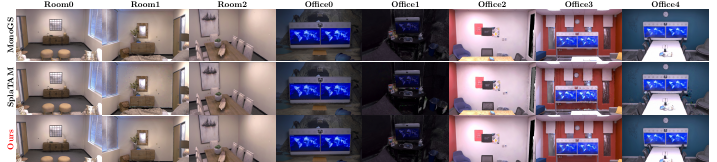}
    \captionsetup{labelfont=bf}
    \caption{
    \textbf{Render Performance in Replica.}
    }
    \vspace{-0.05in}
    \label{fig:render_replica}
\end{figure*}

\begin{figure}[htbp]
    \centering
    
    \includegraphics[width=\columnwidth]{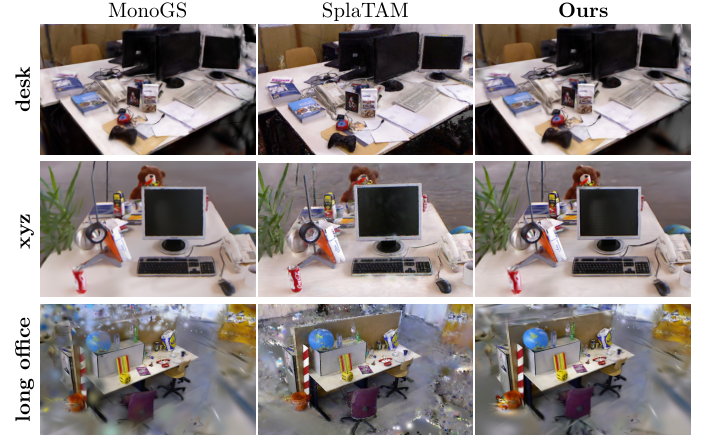}
    \captionsetup{labelfont=bf}
    \caption{
    \textbf{Render Performance in TUM.}
    }
    \vspace{-0.05in}
    \label{fig:render_tum}
\end{figure}

Table \ref{tab:render} shows the rendering performance of MotionGS on the Replica dataset, compared with NICE-SLAM, Point-SLAM, MonoGS,  SplaTAM and GS-SLAM. The result  shows that 3DGS-based SLAM has competitive rendering performance compared with the state-of-the-art NeRF-based SLAM.
Moreover, MotionGS attains the best PSNR and LPIPS on each scenes, outperforming the current state-of-the-art 3DGS-based SLAM. Although the SSIM metrics of MotionGS are not the optimal performance on each scenes, the mean of all SSIM metrics ranks first. We attribute the success to the mask on the scale and opacity of Gaussian, which protects MotionGS from the influence of the almost transparent or negligible Gaussians in rendering. The scene visualizations in these datasets are showed in the Figs \ref{fig:render_tum} and \ref{fig:render_replica}. Compared with the baselines, higher fidelity and more scene details are achieved in the MotionGS.  Our rendering effect showcases enhanced detail representation, as exemplified by the clocks, flower pots, murals, and books within the scene.

\begin{table*}[htbp]
\centering

\begin{tabular}{ccccccccccc}
\toprule
\textbf{Method} & \textbf{Metric} & \textbf{r0} & \textbf{r1} & \textbf{r2} & \textbf{o0} & \textbf{o1} & \textbf{o2} & \textbf{o3} & \textbf{o4} & \textbf{Arg} \\
\midrule

\multirow{3}*{NICE-SLAM} & PSNR$\left [ \mathrm{dB} \right ] \uparrow$ & 22.12 & 22.47 & 24.52 & 29.07 & 30.34 & 19.66 & 22.23 & 24.94 & 24.42\\
~ & $\mathrm{SSIM}\uparrow$ & 0.689 & 0.757 & 0.814 & 0.874 & 0.886 & 0.797 & 0.801 & 0.856 & 0.809\\
~ & $\mathrm{LPIPS}\downarrow$ & 0.33 & 0.271 & 0.208 & 0.229 & 0.181 & 0.235 & 0.209 & 0.198 & 0.233\\
\midrule
\multirow{3}*{Point-SLAM} & $\mathrm{PSNR}\left [ \mathrm{dB} \right ] \uparrow$ & 32.40 & \textcolor[HTML]{009901}{\textbf{34.08}} & \textcolor[HTML]{009901}{\textbf{35.50}} & \textcolor[HTML]{009901}{\textbf{38.26}} & 39.16 & \textcolor[HTML]{009901}{\textbf{33.99}} & \textcolor[HTML]{009901}{\textbf{33.48}} & \textcolor[HTML]{009901}{\textbf{33.49}} & \textcolor[HTML]{009901}{\textbf{35.17}}\\
~ & $\mathrm{SSIM}\uparrow$ &\textcolor{blue}{\textbf{0.974}} & \textcolor{red}{\textbf{0.977}} & \textcolor{red}{\textbf{0.982}} & \textcolor[HTML]{009901}{\textbf{0.983}} & \textcolor{blue}{\textbf{0.986}} & 0.960 & 0.960 & \textcolor{red}{\textbf{0.979}} & \textcolor{blue}{\textbf{0.975}}\\
~ & $\mathrm{LPIPS}\downarrow$ & 0.113 & 0.116 & 0.111 & 0.100 & 0.118 & 0.156 & 0.132 & 0.142 & 0.124\\

\midrule
\multirow{3}*{MonoGS} & $\mathrm{PSNR}\left [ \mathrm{dB} \right ] \uparrow$ & \textcolor{blue}{\textbf{34.83}} & \textcolor{blue}{\textbf{36.43}} & \textcolor{blue}{\textbf{37.49}} & \textcolor{blue}{\textbf{39.95}} & \textcolor{blue}{\textbf{42.09}} & \textcolor{blue}{\textbf{36.24}} & \textcolor{blue}{\textbf{36.7}} & \textcolor{blue}{\textbf{36.07}} & \textcolor{blue}{\textbf{37.50}}\\
~ & $\mathrm{SSIM}\uparrow$ & 0.954 & 0.959 & 0.965 & 0.971 & 0.977 & 0.964 & \textcolor[HTML]{009901}{\textbf{0.963}} & 0.957 & 0.960\\
~ & $\mathrm{LPIPS}\downarrow$ & \textcolor{blue}{\textbf{0.068}} &\textcolor[HTML]{009901}{\textbf{0.076}} & \textcolor{blue}{\textbf{0.075}} & \textcolor{blue}{\textbf{0.072}} & \textcolor[HTML]{009901}{\textbf{0.055}} & \textcolor{blue}{\textbf{0.078}} & \textcolor{blue}{\textbf{0.065}} & \textcolor{blue}{\textbf{0.099}} & \textcolor{blue}{\textbf{0.070}}\\
\midrule
\multirow{3}*{SplaTAM} & $\mathrm{PSNR}\left [ \mathrm{dB} \right ] \uparrow$ &  \textcolor[HTML]{009901}{\textbf{32.86}}& 33.89& 35.25& \textcolor[HTML]{009901}{\textbf{38.26}}& \textcolor[HTML]{009901}{\textbf{39.17}}& 31.97& 29.70& 31.81 &34.11\\
~ & $\mathrm{SSIM}\uparrow$ & \textcolor{red}{\textbf{0.98}}& \textcolor[HTML]{009901}{\textbf{0.97}}& \textcolor{blue}{\textbf{0.98}}& 0.98 & \textcolor[HTML]{009901}{\textbf{0.98}}& \textcolor[HTML]{009901}{\textbf{0.97}}& 0.95& 0.95& \textcolor[HTML]{009901}{\textbf{0.97}}\\
~ & $\mathrm{LPIPS}\downarrow$ & \textcolor[HTML]{009901}{\textbf{0.07}}& 0.10 & \textcolor[HTML]{009901}{\textbf{0.08}}& \textcolor[HTML]{009901}{\textbf{0.09}}& 0.09& 0.10 & 0.12 & 0.15& 0.10\\
\midrule
\multirow{3}*{GS-SLAM} & $\mathrm{PSNR}\left [ \mathrm{dB} \right ]\uparrow $ &  31.56 & 32.86 &32.59& 38.70& \textcolor[HTML]{009901}{\textbf{41.17}}& 32.36& 32.03& 32.92& 34.27\\
~ & $\mathrm{SSIM}\uparrow$ & \textcolor[HTML]{009901}{\textbf{0.968}}& \textcolor{blue}{\textbf{0.973}}& 0.971 & \textcolor{red}{\textbf{0.986}}& \textcolor{red}{\textbf{0.993}} &\textcolor{red}{\textbf{0.978}}& \textcolor{blue}{\textbf{0.97}} & \textcolor[HTML]{009901}{\textbf{0.968}} & \textcolor{blue}{\textbf{0.975}}\\
~ & $\mathrm{LPIPS}\downarrow$ & 0.094& \textcolor{blue}{\textbf{0.075}} &0.093& \textcolor{red}{\textbf{0.050}}& \textcolor{blue}{\textbf{0.033}} & \textcolor[HTML]{009901}{\textbf{0.094}} & \textcolor[HTML]{009901}{\textbf{0.110}}& \textcolor[HTML]{009901}{\textbf{0.112}} &\textcolor[HTML]{009901}{\textbf{0.082}}\\
\midrule
\multirow{3}*{\textbf{Ours}} & $\mathrm{PSNR}\left [ \mathrm{dB} \right ] \uparrow$ & \textcolor{red}{\textbf{36.58}} & \textcolor{red}{\textbf{38.82}} & \textcolor{red}{\textbf{39.71}} & \textcolor{red}{\textbf{43.47}} & \textcolor{red}{\textbf{43.85}} & \textcolor{red}{\textbf{38.87}} & \textcolor{red}{\textbf{37.66}} & \textcolor{red}{\textbf{37.86}} & \textcolor{red}{\textbf{39.60}}\\
~ & $\mathrm{SSIM}\uparrow$ & \textbf{0.967} & \textcolor{blue}{\textbf{0.973}} & \textcolor[HTML]{009901}{\textbf{0.977}} & \textcolor{blue}{\textbf{0.985}} & \textcolor[HTML]{009901}{\textbf{0.985}} & \textcolor{blue}{\textbf{0.976}} & \textcolor{red}{\textbf{0.973}} & \textcolor{blue}{\textbf{0.97}} & \textcolor{red}{\textbf{0.976}}\\
~ & $\mathrm{LPIPS}\downarrow$ & \textcolor{red}{\textbf{0.05}} & \textcolor{red}{\textbf{0.05}} & \textcolor{red}{\textbf{0.05}} & \textcolor{red}{\textbf{0.05}} & \textcolor{red}{\textbf{0.03}} & \textcolor{red}{\textbf{0.042}} & \textcolor{red}{\textbf{0.041}} & \textcolor{red}{\textbf{0.031}} & \textcolor{red}{\textbf{0.043}}\\
\bottomrule
\end{tabular}

\captionsetup{labelfont=bf}
\caption{
\textbf{Comparison of rendering results on Replica.} The red, blue, and green in the above table represent the first, second, and third, respectively.
}
\vspace{-0.05in}
\label{tab:render}
\end{table*}

\subsubsection{Ablation Analysis}
The ablation experiments are showed in Tables \ref{tab:ablation} and \ref{tab:Storage}. It is clearly evident that keyframe selection exerts a profound influence on the tracking accuracy of MotionGS, thereby highlighting the advantage of the dual keyframe selection strategy. 
Furthermore, the SplaTAM has nearly 0.5GB of storage memory as shown in Table \ref{tab:Storage}, while storage memory of MotionGS is less than 50MB. The mask effectively reduces the number of Gaussians utilized in the scene representation, making it a more efficient solution in comparison with the SplaTAM.
\begin{table}[htbp]
\centering
\scalebox{1}{%

\begin{tabular}{cccccc}
\toprule
\textbf{Method (RGB-D)} & \textbf{Metric} &\textbf{fr1} & \textbf{fr2} &\textbf{ fr3} &\textbf{Arg} \\
\midrule
Out of kf strategy & ATE $\downarrow$ &1.87 & 1.27 & 3.48 & 2.21 \\
\textbf{Our} & ATE$\downarrow$ &\textbf{1.47} & \textbf{1.38}&\textbf{1.41} &\textbf{1.46} \\
\bottomrule
\end{tabular}

}
\captionsetup{labelfont=bf}
\caption{
\textbf{Ablation Analysis of the KF Strategy} 
}
\vspace{-0.05in}
\label{tab:ablation}
\end{table}

\begin{table}[htbp]
\centering
\resizebox{\columnwidth}{!}{%

\begin{tabular}{cccccccccc}
\toprule
\textbf{Method} & \textbf{Metric}& \textbf{r0} & \textbf{r1} &\textbf{r2} &\textbf{o0} &\textbf{o1} &\textbf{o2} &\textbf{o3} &\textbf{o4} \\
\midrule
\multirow{2}*{SplaTAM} & Storage(Mb) & 329 & 436 & 371 & 404 & 365 & 295 & 341 & 346\\
~ & Num of Gaussian($10^{4}$) & 509 & 673 & 573 & 624 & 564 & 455 & 527 & 534 \\
\multirow{2}*{\textbf{Ours$^{*}$}} & Storage(Mb) & 20.5 & 18.8 & 18.3 & 18.4 & 13.8 & 26.0 & 22.2 & 20.5\\
~ & Num of Gaussian($10^{4}$) & 37 & 29 & 28 & 28 & 21 & 39 & 36 & 32 \\
\multirow{2}*{\textbf{Ours}} & Storage(Mb) & \textbf{16.6} & \textbf{14.2} & \textbf{15.9} & \textbf{15.6}&  \textbf{11.8} & \textbf{23.6} & \textbf{20.8} & \textbf{17.6}\\
~ & Num of Gaussian($10^{4}$) & \textbf{31} & \textbf{24} & \textbf{26} & \textbf{25} & \textbf{18} & \textbf{35} & \textbf{33} & \textbf{27} \\

\bottomrule
\end{tabular}
}
\captionsetup{labelfont=bf}
\caption{
\textbf{Ablation Analysis of the Mask loss}. 
 The Ours$^{*}$ refers to the MotionGS without mask loss.
}
\vspace{-0.05in}
\label{tab:Storage}
\end{table}


\section{Conclusion}
\label{sec:conclusion}
This study proposed a 3DGS-based SLAM named MotionGS that integrates deep visual features, dual keyframe selection, and 3DGS. With its exquisite design, the state-of-the-art performance of the MonoGS has been fully demonstrated in extensive experiments. The proposed approach further emphasizes the broad potential of 3DGS in the SLAM field. Based on this work, the multi-sensor 3DGS-based SLAM tailored for large-scale outdoor scenes would be the next research.


{\small
\bibliographystyle{IEEEtran}
\bibliography{11_references}
}

\end{document}


\title{\paperTitle}
\author{\authorBlock}
\maketitlesupplementary

\appendix
\section{Appendix Section}
\label{sec:appendix_section}
Supplementary material goes here.

{\small
\bibliographystyle{ieeenat_fullname}
\bibliography{11_references}
}